# A Multi-Agent Simulation of Retail Management Practices

**Peer-Olaf Siebers, Uwe Aickelin**
**Automated Scheduling, Optimisation and Planning Research Group, School of Computer Science & IT, University of Nottingham, Nottingham, NG8 1BB, UK**
pos@cs.nott.ac.uk, uxa@cs.nott.ac.uk

**Helen Celia, Chris Clegg**
**Centre for Organisational Strategy, Learning & Change, Leeds University Business School, University of Leeds, Leeds, LS2 9JT, UK**
h.celia@leeds.ac.uk, c.w.clegg@leeds.ac.uk

**Abstract**

We apply Agent-Based Modeling and Simulation (ABMS) to investigate a set of problems in a retail context. Specifically, we are working to understand the relationship between human resource management practices and retail productivity. Despite the fact we are working within a relatively novel and complex domain, it is clear that intelligent agents do offer potential for developing organizational capabilities in the future. Our multi-disciplinary research team has worked with a UK department store to collect data and capture perceptions about operations from actors within departments. Based on this case study work, we have built a simulator that we present in this paper. We then use the simulator to gather empirical evidence regarding two specific management practices: empowerment and employee development.

## 1. INTRODUCTION

It is well-documented that the UK's productivity levels tend to lag behind those of countries with comparably developed economies [1]. The retail sector in particular has been identified as one of the biggest contributors to the productivity gap that persists between the UK, other European countries and the USA [2].

There is no doubt that management practices are linked to productivity and the performance of a company [3]. Best practices have been developed, but when it comes down to the actual application of these guidelines considerable ambiguity remains regarding their effectiveness in a particular scenario [4].

Operational Research (OR) is a discipline that applies advanced analytical methods to help make better informed decisions. It is used for problems concerning the conduct and co-ordination of the operations within an organization [5]. An OR study usually involves the development of a scientific model that attempts to abstract the essence of the real problem. When investigating the behavior of complex systems the choice of an appropriate modeling technique is very important.

Most OR methods can only be used as analysis tools once management practices have been implemented. Often they are not very useful for giving answers to speculative 'what-if' questions, particularly when one is interested in the development of the system over time rather than just a snapshot.

Simulation is an OR method that can be used to analyze the operation of dynamic and stochastic systems. ABMS is particularly useful when complex interactions exist between system entities, for example the processes of autonomous decision-making or negotiation. In ABMS the researcher explicitly describes the decision process of simulated actors at the micro level. Structures emerge at the macro level as a result of the actions of the individual agents, interactions between agents, and also with their environment.

ABMS offers a new and exciting way of understanding the world of work and hence ABMS and its application to management practices carries great potential. We have developed simulation models based on research by our multi-disciplinary team of economists, work psychologists and computer scientists.

In this paper we show how agent-based simulation experiments can deal with assessing and optimizing management practices such as training, empowerment or teamwork. We will discuss the experience we have gained whilst implementing these concepts within a well-known retail department store.

## 2. WHY AGENT-BASED SIMULATION?

Currently there is no reliable and valid way to wholly delineate the effects of management practices from other socially embedded factors. Our current work hones in on the UK retail sector, but what we are learning about system modeling has implications for modeling any complex system that involves many human interactions and where the actors work with some degree of autonomy.

A recent literature review [4] reveals that previous research into retail productivity has typically focused on consumer behavior and efficiency evaluation. We seek to build on this work and address the neglected area of retail management practices [6].

In terms of commercial software, ShopSim [7] is an example of a decision support tool designed for retail and shopping centre management. It uses an agent-based approach, where behavior of agents is directed by survey data. However, the software only evaluates the layout and design of a shopping centre and does not allow investigation of the effectiveness of certain management practices.

In summary, we can say that only limited work has been conducted into the development of models that would allow an investigation of the impact of management practices on retail productivity.

In order to select an appropriate modeling technique, we reviewed the relevant literature spanning the fields of Economics, Social Science, Psychology, Retail, Marketing, OR, Artificial Intelligence, and Computer Science. Within these fields a wide variety of approaches are used which can be classified into three main categories: analytical approaches, heuristic approaches, and simulation. In many cases we found that combinations of these were used within a single model ([8; 9]). From these approaches we identified simulation as best suiting our needs.

Simulation introduces the possibility of a new way of thinking about social and economic processes, based on ideas about the emergence of complex behavior from relatively simple activities [10]. This modeling technique allows clarification, implementation, and validation of a theory. While analytical models typically aim to explain correlations between variables measured at one single point in time, simulation models are concerned with the development of a system over time. Furthermore, analytical models tend to operate on a much higher level of abstraction than simulation models.

The effectiveness of a simulation model depends upon the right level of abstraction. Csik [11] states that on the one hand the number of free parameters should be kept as low as possible. On the other hand, too much abstraction and simplification might threaten the fit between reality and the breadth of the simulation model. There are several different paradigms existing in simulation modeling. The major ones are Discrete Event (DE), System Dynamics (SD), and Agent Based (AB) [12]. The choice of the most suitable approach always depends on the issues investigated, the input data available, the level of analysis and the type of answers that are sought. Technically, SD deals mostly with continuous processes whereas DE and AB operate mostly in discrete time steps.

Although computer simulation has been used widely since the 1960s, ABMS only became popular in the early 1990s [13]. It is described by Jeffrey [14] as a mindset as much as a technology: 'It is the perfect way to view things and understand them by the behavior of their smallest components'. ABMS can be used to study how micro-level processes affect macro level outcomes. A complex system is represented by a collection of agents that are programmed to follow simple behavioral rules. Agents can interact with each other and with their environment to produce complex collective behavioral patterns. Macro behavior is not explicitly modeled; it emerges from the micro-decisions of individual agents [15].

The main characteristics of agents are their autonomy, their ability to take flexible action in reaction to their environment and their pro-activeness depending on motivations generated from their internal states. They are designed to mimic the attributes and behaviors of their real-world counterparts. The simulation output may be used for explanatory, exploratory and predictive purposes [16]. This approach offers a new opportunity to realistically and validly model organizational characters and their interactions, to allow a meaningful investigation of management practices. ABMS is still a relatively new simulation technology and its principal application has been in academic research. With the availability of more sophisticated modeling tools, things are starting to change [17]. In addition ABMS is extensively used by the game and film industry to develop realistic simulations of individual characters and societies. It is used in computer games, for example The SIMS™ [18], or in films when diverse heterogeneous characters animations are required, for example the Orcs in Lord of the Rings™ [19].

Due to the characteristics of the agents, this modeling approach appears to be more suitable than the DE one for modeling human-based systems. ABMS seems to promote a natural form of modeling, as active entities in the live environment are interpreted as actors in the model. There is a structural correspondence between the real system and the model representation, which makes them more intuitive and easier to understand than for example a system of differential equations as used in SD.

Hood [20] emphasizes that one of the key strengths of ABMS is that the system as a whole is not constrained to exhibit any particular behavior as the system properties emerge from its constituent agent interactions. Hence, assumptions of linearity, equilibrium and so on, are not needed. On the other hand, there is consensus in the literature that it is difficult to evaluate agent-based models, because the behavior of the system emerges from the interactions between the individual entities. Furthermore, problems often occur through the lack of adequate real data.

## 3. MODEL DESIGN AND IMPLEMENTATION

### 3.1. Model Concepts

Case studies were undertaken in four departments across two retail stores in a single company. The studies involved extensive data collection spanning: participant observation, semi-structured interviews with team members, management and personnel, completion of survey questionnaires and the analysis of company data and reports. Research findings were consolidated and fed back (via report and presentation) to employees with extensive experience and knowledge of the four departments in order to validate our understanding and conclusions. This approach has enabled us to acquire a valid and reliable understanding of how the real system operates, revealing insights into the working of the system as well as the behavior of and interactions between the different agents within it.

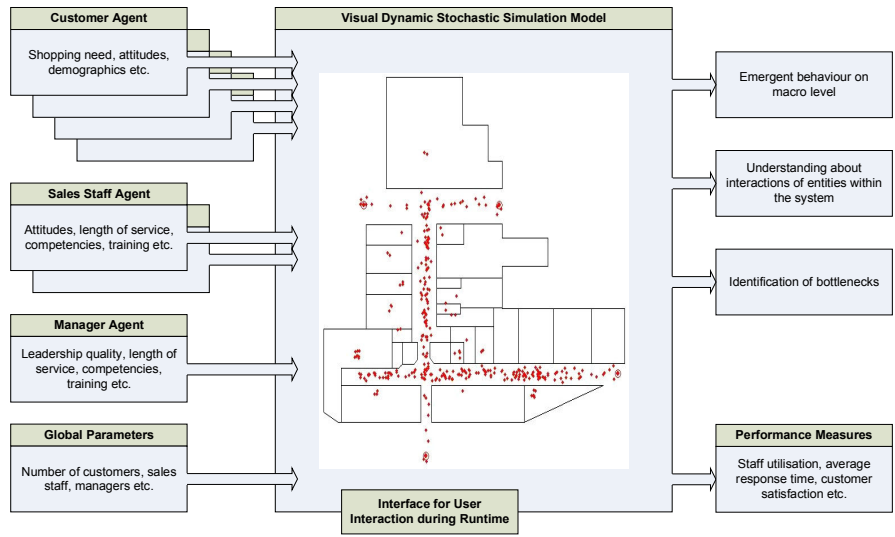

**Figure 1.** Conceptual model for the simulator

Our initial ideas for the simulator are shown in Figure 1. Within our conceptual model we have three different types of agents (customers, sales staff and managers), each with a different set of attributes. We use probabilities and frequency distributions to assign different values to each individual agent. In this way a heterogeneous population is created that reflects the variations in attitudes and behaviors of their real human counterparts. In addition, we need to incorporate global parameters such as the number of agents. Regarding system outputs, we aim to find some emergent behavior on a macro level. Visual representation of the simulated system and its actors allows us to monitor and better understand the interactions of entities within the system. Coupled with standard performance measures (e.g. utilization) we aim to identify bottlenecks to assist with optimization of the modeled system.

### 3.2. Agent Design

Our agents are conceptualized in state charts. State charts show the different states an entity can be in, and define possible transitions from one state to another and the events that cause them. This is exactly the information we need in order to represent our agents within the simulation environment. Also, this form of graphical representation is helpful for validating the agent design, as non-specialists can easily understand it.

The art of modeling is simplification and abstraction [21]. A model is always a restricted copy of the real world. Researchers have to identify the most important components of a system to build effective models. In our case, instead of looking for components, we have to identify the most important behaviors of an actor and the triggers that initiate a move from one state to another; for example when a certain period of time has elapsed, or at the request of another agent. We have developed state charts for all of the agents in our model. Figure 2 shows one of the state charts, in this case for a customer agent.

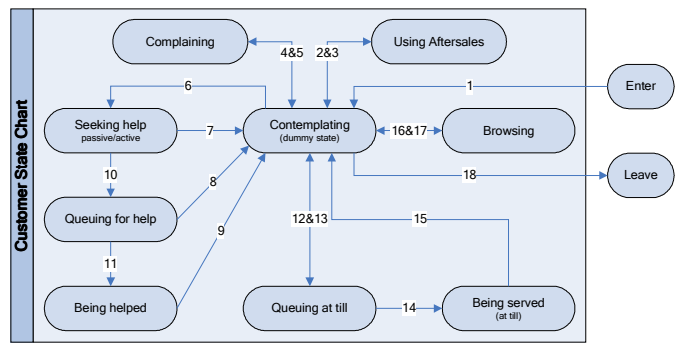

**Figure 2.** Conceptual model for customer agent (transition rules have been omitted for simplicity)

A customer enters the department in the contemplating state. This is a dummy state and represents the reality of an individual thinking through their behavioral intentions prior to acting [22], whether a planned or unanticipated purchase [23]. Even when a particular purchase is planned, the consumer may change their mind and go for a substitute product, if they buy at all. S/he will probably start browsing and after a certain amount of time (delay value derived from a probability distribution) s/he may require help, queue at the till or leave the shop. If the customer requires help, s/he considers what to do and seeks help by sending a message to a staff member and will either immediately receive help or wait for attention.. If no staff member is available, s/he has to wait (queue) for help. Whilst waiting, s/he may browse for another item, proceed to the till to buy a chosen item, or may leave the shop prematurely if the wait is too long.

### 3.3. Implementation

Our simulation has been implemented in AnyLogic™, which is a Java™ based multi-paradigm simulation software [24]. Currently the simulator can represent the following actors: customers, service staff (including cashiers and selling staff of two different training levels) and section managers. Figure 3 shows a screenshot of the current customer and staff agent logic as it has been implemented in AnyLogic™. Boxes show the possible states, arrows the possible transitions and numbers reveal satisfaction weights.

The customer agent template consists of three main blocks which use a very similar logic. In each block, in the first instance, a customer will try to obtain service directly and if s/he cannot obtain it (no suitable staff member is available) s/he will have to queue. The customer will then either be served as soon as an appropriate staff member becomes available, or they will leave the queue if they do not want to wait any longer (an autonomous decision). A complex queuing system has been implemented to support different queuing rules. The staff agent template, in comparison to the customer agent template, is relatively simple. Whenever a customer requests service and the staff member is available and has the right level of expertise for the task requested, the staff member commences this activity until the customer releases the staff member.

We introduce a service level index as a novel performance measure using the satisfaction weights mentioned earlier. This index allows customer service satisfaction to be recorded throughout the simulated lifetime. The idea is that certain situations exert a bigger impact on customer satisfaction than others, and we can assign weights to events to account for this. This helps the analyst to find out to what extent customers had a positive or negative shopping experience. It also allows the analyst to put emphasis on different operational aspects of the system, and try out the impact of different strategies.

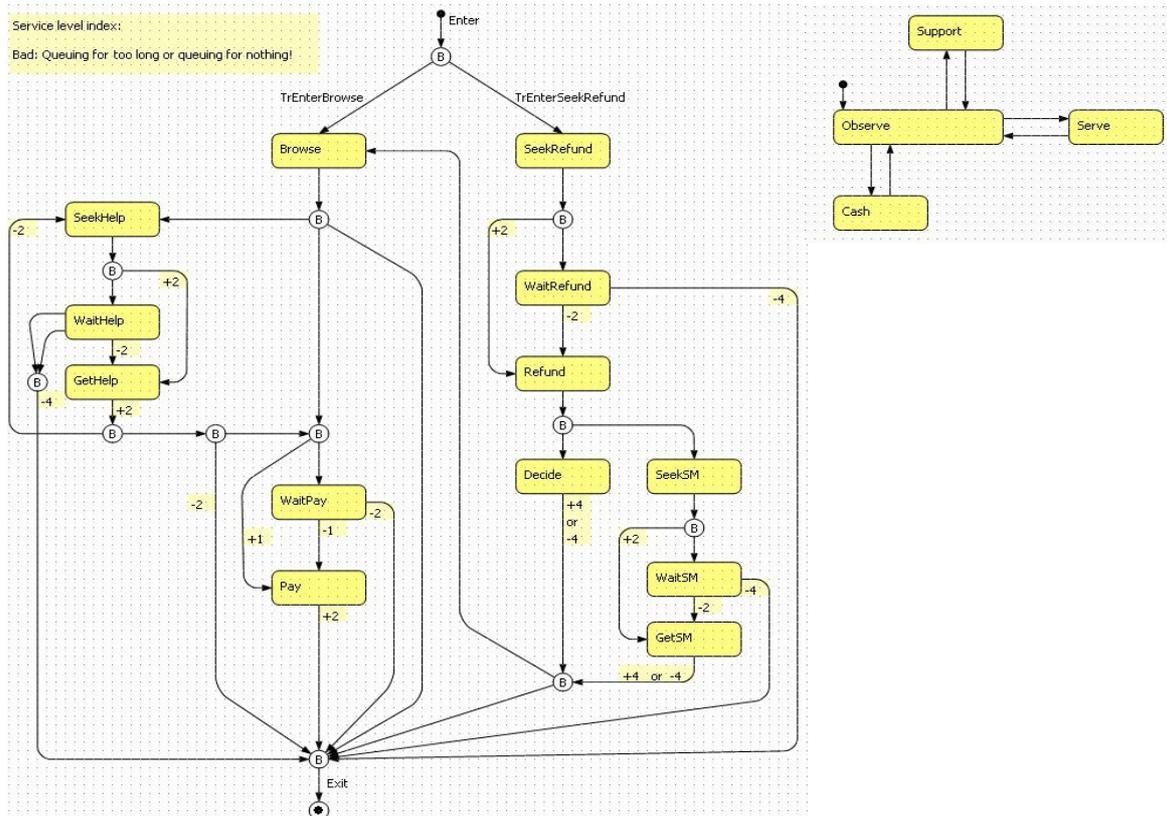

**Figure 3.** Customer (left) and staff (right) agent logic implementation in AnyLogic™

Currently the simulator supports the simulation of the two department types we looked at during the case study: Womenswear (WW) and Audio & Television (A&TV). These department types differ with respect to their operational structure, staff composition, service provision and customer types. WW customers will ask for help when they know what they want whereas A&TV customers will ask for help when they do not know what they want. WW makes a lot more unassisted sales than A&TV and service times are very different. In WW, the average service time is a lot shorter than in A&TV, and the average price of the items sold assisted in A&TV is a lot higher.

The data used in the simulator is partly real data collected during the case study, partly estimates gathered during interviews and partly estimates from observations. Collecting numerical data has been difficult as not a lot of the operational data needed for our simulation is gathered by the case study company and the available data (mainly performance data) is often in an inappropriate format. For example, different departments are combined or averaged over different time periods.

## 4. A FIRST VALIDATION OF THE SIMULATOR

To test the operation of our simulator and establish its validity we have designed and run 3 sets of experiments. We investigate the impact of two management practices: empowerment and employee development. The staff group in every experiment consisted of 3 cashiers, 7 normal selling staff and 2 experts, with a customer arrival rate of 70 per hour, and a runtime of 10 weeks. We have systematically manipulated only the independent variable of interest in each experiment. We have conducted at least 20 replications for every experimental condition enabling the application of rigorous statistical techniques.

Each set of results was analyzed using a one-way between-groups analysis of variance (ANOVA). Despite our prior knowledge of how the real system operates, we were unable to hypothesize precise differences (for example, turning points) in variable relationships, instead predicting general patterns of relationships. Indeed, ABMS is a decision-support tool and is only able to inform us about directional changes between variables (actual figures are notional). Where significant ANOVA results were found, post-hoc tests were applied to investigate further the precise impact on outcome variables under different experimental conditions. To address the increased risk of a Type I error we have applied a Bernoulli correction to create more conservative thresholds for significance (corrected post-hoc p-value for 3 dependent variables = .0167).

During our case study work, we observed the implementation of a new refund policy. This new policy allows any cashier to decide independently whether to make a customer refund up to the value of £50, rather than being required to refer the authorization to an expert employee. To first simulate the impact of this practice on key business outcomes, we have systematically varied the probability that employees are empowered to make refund decisions autonomously. Cashiers were configured to process a refund in 80% of cases, whereas experts were more critical and only accept 70% of refund claims.

As we increase the level of empowerment, we expect to see more transactions as work flows more effectively and cashiers can take more decisions autonomously and quickly without requiring expert assistance. We also anticipate greater levels of customer satisfaction (whether obtaining a refund or not), because staff time is less consumed by the delays of locating expert assistance, resulting in more employee time available to customers. As the level of empowerment increases, we predict:

- H1. higher numbers of transactions.
- H2. greater customer satisfaction
- H3. higher refund satisfaction

An ANOVA revealed statistically significant differences across all three outcomes: number of transactions [$F(4, 95)=26.77$, $p<.01$], customer satisfaction [$F(4, 95)=12.35$, $p<.01$], and refund satisfaction [$F(4, 95)=2001.73$, $p<.01$]. Consulting Table 1, we see that H1 has not been supported, and the number of transactions actually decreases with empowerment, whereas H2 and H3 are confirmed. The effect size, calculated using eta squared, reveals differences in the relative impact of empowerment on each outcome measure: 0.53 for the number of transactions, 0.34 for customer satisfaction and 1.00 for refund satisfaction. Social scientists report 0.14 as indicative of a large effect [25] suggesting we are looking at substantial effect sizes.

Post-hoc comparisons using the Tukey's test indicated a number of significant differences between group means. Most notably the impact on refund satisfaction was huge, with every single increment in empowerment resulting in a significant increase in refund satisfaction. H1 was not supported. This unforeseen reduction in transactions may have occurred because less experienced employees take longer to make a decision on a refund, resulting in a knock-on impact for customer waiting times. H2 holds, and this finding is intuitive because customers prefer that one staff member can deal with their needs. H3 is strongly supported, and makes sense because cashiers are also more likely to approve a customer refund request. In reality, we also know that customers generally prefer to deal with a single customer representative.

**Table 1.** Descriptive statistics for Experiment 1: Empowerment outcome variables (all to two d.p.)

| Empower-ment level | Number of Transactions | | Overall Satisfaction | | Refund Satisfaction | |
|---|---|---|---|---|---|---|
| | Mean | S.D. | Mean | S.D. | Mean | S.D. |
| 0 | 15133.85 | 102.02 | 23554.30 | 892.55 | -3951.40 | 288.84 |
| 0.25 | 15114.75 | 60.04 | 24331.35 | 907.02 | -2316.10 | 187.23 |
| 0.5 | 15078.95 | 86.24 | 24476.95 | 907.48 | -932.40 | 243.25 |
| 0.75 | 15008.45 | 52.53 | 25213.10 | 898.61 | 613.70 | 182.03 |
| 1 | 14920.15 | 66.42 | 25398.95 | 1092.50 | 1892.80 | 237.69 |

Our case study work has revealed that a key way in which employees can develop their product knowledge occurs when they are unable to fully meet a customer's request for advice. An expert is called over and the original employee is empowered to choose whether or not to stay with them to learn from the interaction. In this second set of experiments we are assuming that, given the opportunity to choose to learn, an employee will usually decide to take up that opportunity. We found that case study employees enjoyed providing excellent customer service, and given the opportunity would do what they could to stay abreast of product developments.

In our model, a normal staff member gains knowledge points on every occasion that he or she stays with an expert. We have systematically varied the probability that a normal staff member learns in this way. Of course, there is a trade-off with short-term ability to meet customer demand, and a customer may leave prematurely if they have to wait for too long. Normal staff members will be occupied for longer when their will to learn is stronger.

By allowing employees to acquire new product knowledge from expert colleagues, we anticipate performance improvements. We predict that increasingly empowering employees to learn will result in:

- H4. an increase in the knowledge of normal staff.
- H5. an increase in the utilization of normal staff.
- H6. no change to the utilization of expert staff.
- H7. a short term reduction in the number of sales transactions.
- H8. a reduction in customer satisfaction.

The second ANOVA (see Table 2 for descriptives) exposed a significant impact of empowerment to learn on: normal staff expertise [$F(4, 96)=2,794.12, p<.01$], utilization of normal staff [$F(4, 96)=112.53, p<.01$], and customer satisfaction [$F(4, 96)=29.16, p<.01$]. Tests of expert staff utilization [$F(4, 96)=1.28, p=.29$] and sales transactions [$F(4, 96)=1.25, p=.30$] were insignificant. Effect sizes of significant relationships were all large (normal staff expertise = 0.99, normal staff utilization = 0.83, customer satisfaction = 0.55).

Tukey's post-hoc comparisons were run for the three significant findings. Both normal staff expertise and utilization significantly increased with every single increment in employee empowerment to learn. The largest, significant differences in overall customer satisfaction are observed at the polar ends of the scale. As predicted, employees who are empowered to learn become more knowledgeable (H4), leading to a more efficient utilization of employees as a whole (H5). H6 holds as expected, meaning there is no significant impact on the utilization of expert staff in terms of the time they spend engaged with customers. However we can see through the effects on other outcome measures that higher levels of learning empowerment result in better 'utilization' of experts; the harnessing of their knowledge. H7 has not been supported as the number of transactions does not significantly differ between experimental conditions. The short-lived reduction that we anticipated appears to be so negligible that it is inconsequential; nonetheless, the associated increase in customer waiting times has negatively influenced the customer service index, providing support for H8.

Our third and final set of experiments goes one step further and explores how time invested in learning impacts on medium-term system performance. Our model mimics an evolutionary process whereby staff members can progressively develop their product knowledge over a period of time. When a staff member has accumulated a certain number of knowledge points from observing expert service transactions, they are considered an expert.

We have systematically varied the number of knowledge points required to attain expert-level competence. All normal staff members are programmed to take advantage of all learning opportunities.

By investing time in developing and expanding employees' specialist knowledge, we anticipate even greater future savings in terms of key outcomes, beyond those already observed in Experiment 2. The academic literature echoes the positive business impact of employing individuals with greater expertise to provide better customer service and advice (e.g. [26]). We predict that increasing the rate of employee development (by lowering the threshold for attaining expert status) will result in more desirable outcome variables, specifically increases in:

- H9. normal staff member expertise.
- H10. normal staff utilization.
- H11. expert staff utilization.
- H12. the number of transactions.
- H13. the customer satisfaction index.

**Table 2.** Descriptive statistics for Experiment 2: Empowerment to learn outcome variables (all to two d.p.)

| Empower-ment to Learn | Normal Expertise | | Utilisation of Normal Staff | | Utilisation of Expert Staff | | Number of Transactions | | Overall Satisfaction | |
|---|---|---|---|---|---|---|---|---|---|---|
| | Mean | S.D. | Mean | S.D. | Mean | S.D. | Mean | S.D. | Mean | S.D. |
| 0 | 0.00 | 0.00 | 0.82 | 0.01 | 0.93 | 0.00 | 14830.00 | 99.82 | 28004.00 | 823.19 |
| 0.25 | 18.36 | 2.10 | 0.83 | 0.01 | 0.94 | 0.00 | 14801.00 | 73.56 | 26937.00 | 960.37 |
| 0.5 | 35.66 | 2.54 | 0.84 | 0.01 | 0.94 | 0.00 | 14782.00 | 79.90 | 26310.00 | 916.38 |
| 0.75 | 53.44 | 2.98 | 0.85 | 0.01 | 0.94 | 0.01 | 14787.00 | 96.45 | 25678.00 | 1269.68 |
| 1 | 69.35 | 2.85 | 0.85 | 0.01 | 0.94 | 0.00 | 14823.00 | 80.42 | 24831.00 | 1043.79 |

**Table 3.** Descriptive statistics for Experiment 3: Learning outcome variables (all to two d.p.)

| Competence threshold | Normal Staff Member | | Normal Staff Utilisation | | Expert Staff Utilisation | | Number of Transactions | | Overall Customer Satisfaction | |
|---|---|---|---|---|---|---|---|---|---|---|
| | Mean | S.D. | Mean | S.D. | Mean | S.D. | Mean | S.D. | Mean | S.D. |
| 0 | 0.00 | 0.00 | 0.00 | 0.00 | 0.77 | 0.01 | 15482.35 | 97.66 | 46125.25 | 1099.48 |
| 0.2 | 0.00 | 0.00 | 0.00 | 0.00 | 0.80 | 0.01 | 15302.85 | 75.00 | 40723.95 | 1209.39 |
| 0.4 | 0.00 | 0.00 | 0.00 | 0.00 | 0.82 | 0.01 | 15125.15 | 52.03 | 34992.75 | 1770.02 |
| 0.6 | 0.00 | 0.00 | 0.00 | 0.00 | 0.86 | 0.01 | 14945.30 | 118.41 | 28958.80 | 1460.78 |
| 0.8 | 67.68 | 3.23 | 0.86 | 0.01 | 0.93 | 0.02 | 14801.95 | 92.79 | 24661.75 | 1058.27 |
| 1 | 68.83 | 3.84 | 0.86 | 0.00 | 0.94 | 0.00 | 14827.90 | 76.14 | 24668.80 | 843.84 |

The final ANOVA revealed statistically significant differences in expert utilization $[F(5,114)=952.21, p<.01]$, volume of transactions $[F(5,114)=193.14, p<.01]$ and overall customer satisfaction $[F(5,114)=959.01, p<.01]$. The effect sizes of significant relationships were again all very large (expert staff utilization = 0.98, volume of transactions = 0.89, and customer satisfaction = 0.98). We were unable to adequately test the impact of learning on normal staff expertise (H9) and utilization (H10), because we do not have this data for all experimental conditions (see Table 3: at the lower promotion thresholds, all normal staff have been promoted by the end of the simulation run).

Tukey's post-hoc comparisons, revealed significant differences in every variable for every single increment in the competence threshold, with the exception of between the two upper levels. However, only expert utilization was in the predicted direction. Therefore our evidence was strongly in favor of H11, whereas the exactly the contrary of H12 and H13 have been supported. This is strongly counter-intuitive because we would expect that the greater the number of resulting experts, the greater the availability of top-quality advice to customers. Indeed, it is possible that our simulation run is too short at just ten weeks, and presents only a backward facing view of department performance; i.e., focusing on the time consumed in learning, and not on the time spent sharing their new competence with customers. We are also assuming that staff acquire expertise purely by learning from their colleagues, whereas in reality this would be supported with other sources and forms of learning.

## 5. CONCLUSIONS AND FUTURE DIRECTIONS

In this paper, we present the conceptual design, implementation and operation of a retail department simulator used to understand the impact of management practices on retail productivity. As far as we are aware, this is the first time that researchers have tried to use agent-based approaches to simulate management practices such as employee development and empowerment. Although our simulator uses specific case studies as source of information, we believe that the general model could be adapted to other retail companies and areas of management practices that have high degrees of human interaction.

From what we can conclude from our current analyses, some findings are as hypothesized whereas others are more mixed. Further experimentation is required to explore the model's operation. Early findings indicate that management practices tend to exert a subtle yet significant effect on performance, consistent with our case study findings.

Currently we are developing our agents with the goal of enhancing their intelligence and heterogeneity. We are planning to introduce evolution and stereotypes. Importantly, organizations generally work in environments where they need to adopt long-term strategies, and so we are developing our model to allow us to evaluate how system outcomes evolve over time, just as we observe in the dynamic reality of the system. We plan to investigate shopping experience based on long-term satisfaction scores, with the overall effect of holding a certain reputation or shopping brand. Another interesting aspect we are currently implementing is the introduction of stereotypes. Our case study organization has identified its particular customer stereotypes through market research, and we plan to find out how populations of certain customer types influence sales.

Taking a step back, we believe that researchers should become more involved in this multi-disciplinary kind of work to gain new insights into the behavior of organizations. In our view, the main benefit from adopting this approach is the improved understanding of and debate about a problem domain, and the resulting overt convergence of understanding and agreement about a system's functioning. The very nature of the methods involved forces researchers to be explicit about the rules underlying behavior and to think in new ways about them. As a result, we have brought work psychology and agent-based modeling closer together to form a new and exciting research area.